\documentclass[lettersize,journal]{IEEEtran}
\ifdefined\pdfgentounicode
  \input{glyphtounicode}
\fi

\usepackage[utf8]{inputenc}
\usepackage[T1]{fontenc}
\usepackage{microtype}
\usepackage[dvipsnames,svgnames,x11names]{xcolor}
\usepackage{amsmath,amssymb,amsfonts,bm,dsfont}
\newcommand{\mathbbm}[1]{\mathds{#1}}
\usepackage{graphicx}
\usepackage{algorithmic}
\usepackage{algorithm}
\usepackage[caption=false,font=normalsize,labelfont=sf,textfont=sf]{subfig}
\usepackage{textcomp,verbatim}
\usepackage{array,booktabs,multirow,colortbl,tabularx}
\usepackage{siunitx}
\usepackage{enumitem}
\usepackage{tcolorbox}
\tcbuselibrary{breakable,skins}
\usepackage{placeins}
\usepackage{float}
\usepackage{stfloats}
\usepackage{balance}
\usepackage{cite}
\usepackage{url}
\usepackage{courier}
\usepackage[pagebackref,breaklinks,colorlinks]{hyperref}
\hypersetup{
  citecolor=green!80!black,
  pdftitle={Learning to Pose Problems: Reasoning-Driven and Solver-Adaptive Data Synthesis},
  pdfauthor={Yongxian Wei, Yilin Zhao, Li Shen, Zixuan Hu, Xinrui Chen, Runxi Cheng, Sinan Du, Hao Yu, Dian Li, Chun Yuan},
  pdfkeywords={Data synthesis, large reasoning models, reinforcement learning, problem generation}
}
\usepackage[capitalize,noabbrev]{cleveref}
\newcommand{\BIBdecl}{\tolerance=1000\emergencystretch=3em\relax}

\DeclareUnicodeCharacter{2011}{-}
\setlist[itemize]{leftmargin=*,itemsep=2pt,topsep=3pt,parsep=0pt}
\setlist[enumerate]{leftmargin=*,itemsep=2pt,topsep=3pt,parsep=0pt}
\tcbset{before skip=5pt,after skip=5pt,boxrule=0.5pt,arc=1.5mm,
  boxsep=1mm,left=1.5mm,right=1.5mm,top=1mm,bottom=1mm}
\fnbelowfloat

\crefname{section}{Sec.}{Secs.}
\Crefname{section}{Section}{Sections}
\crefname{table}{Table}{Tables}
\Crefname{table}{Table}{Tables}
\crefname{equation}{Eq.}{Eqs.}
\crefname{algorithm}{Algorithm}{Algorithms}
\crefname{figure}{Fig.}{Figs.}
\crefname{appendix}{Appendix}{Appendices}
\Crefname{appendix}{Appendix}{Appendices}
\newcommand{\ie}{\emph{i.e.}}
\newcommand{\eg}{\emph{e.g.}}

\title{Learning to Pose Problems: Reasoning-Driven and Solver-Adaptive Data Synthesis}

\author{Yongxian~Wei, Yilin~Zhao, Li~Shen, Zixuan~Hu, Xinrui~Chen, Runxi~Cheng,\\
  Sinan~Du, Hao~Yu, Dian~Li, and Chun~Yuan,~\IEEEmembership{Senior Member,~IEEE}%
\thanks{Yongxian~Wei, Xinrui~Chen, Runxi~Cheng, Sinan~Du, Hao~Yu, and Chun~Yuan are with Shenzhen International Graduate School, Tsinghua University, Shenzhen 518071, China (email: weiyx23@mails.tsinghua.edu.cn, cxr22@tsinghua.org.cn, crx23@mails.tsinghua.edu.cn, dsn23@mails.tsinghua.edu.cn, yuh24@mails.tsinghua.edu.cn, yuanc@sz.tsinghua.edu.cn).\par
Yilin~Zhao and Dian~Li are with Tencent (email: yilinnzhao@tencent.com, goodli@tencent.com).\par
Zixuan~Hu is with Nanyang Technological University, Singapore 639798 (email: ZIXUAN014@e.ntu.edu.sg).\par
Li~Shen is with the School of Cyber Science and Technology, Shenzhen Campus of Sun Yat-sen University, Shenzhen 518107, China (email: mathshenli@gmail.com).}}

\begin{document}

\maketitle
\begin{abstract}
Data synthesis for training large reasoning models offers a scalable alternative to limited, human-curated datasets, enabling the creation of high-quality data.
However, existing approaches face several challenges: (i) indiscriminate generation that ignores the solver’s ability and yields low‑value problems, or reliance on complex data pipelines to balance problem difficulty; and (ii) a lack of reasoning in problem generation, leading to shallow problem variants.
In this paper, we develop a problem generator that \textit{reasons explicitly} to plan problem directions before synthesis and \textit{adapts difficulty} to the solver’s ability.
We recover problem-design CoT from human-authored multi-part questions, using later subquestions as targets to infer the design rationale. These traces initialize the generator, which is then optimized with a seed-conditioned reward based on solver answer consistency. The reward guides difficulty adjustment relative to each seed while favoring problems near the solver's competence boundary. Across 10 mathematical and general reasoning benchmarks, our method outperforms seed-only training by 2.69 and 3.39 percentage points after one and three rounds, respectively. The framework also extends to vision-language models. Our code is publicly available at \url{https://github.com/WalkerWorldPeace/Reasoning-Synthesis}.
\end{abstract}


\section{Introduction}\label{sec:intro}

\IEEEPARstart{R}{ecently}, Large Reasoning Models (LRMs) such as OpenAI-o1~\cite{openai2024llmreasoning} and DeepSeek-R1~\cite{guo2025deepseek} have demonstrated enhanced problem-solving capabilities through extended chain-of-thought (CoT). These models are typically trained via reinforcement learning with verifiable rewards (RLVR) on carefully curated datasets. While human-crafted data remains valuable, data synthesis has emerged as a promising direction for advancing LRMs~\cite{jiao2025img,wu2025synthetic}, given the scarcity of curated datasets and the scalability of synthetic pipelines.

Automated synthesis of mathematical problems commonly prompts proprietary or large open‑source models with seed problems~\cite{tang2024mathscale,lu2024mathgenie,luo2025wizardmath}. However, existing methods either rely on indiscriminate generation, which yields low‑value problems, or on complex data pipelines to calibrate difficulty. For example, directly prompting a commercial API-based LLM~\cite{xu2023wizardlm} can produce diverse problems, but does not account for the solver’s ability. In contrast, complex data pipelines calibrate difficulty offline, generating large candidate pools that are subsequently scored and filtered by auxiliary estimators, which incurs substantial compute overhead~\cite{pei2025scalediff}. Furthermore, these methods do not endow the generator with explicit reasoning capability for data synthesis, resulting in shallow variants of the problems. While CoT and RLVR have been widely used to improve problem \emph{solving}, problem \emph{generation} is still largely treated as a direct prompting or filtering process, with limited explicit reasoning about problem design.

Language and multimodal reasoning are commonly strengthened via CoT-based supervised fine-tuning (SFT)~\cite{shi2025enhancing}, causal CoT or logic-integrated inference for visual reasoning~\cite{li2026visual,xue2025linin}, collaborative multimodal inference~\cite{wen2025enhanced}, and RLVR. Yet these methods reason during answer derivation rather than problem design, and transferring RLVR from solving to generation is difficult: multiple valid outputs may exist and no canonical answer is available during generation, making deterministic rule checking impractical. Consequently, practitioners often rely on low‑scalability human annotation or LLM‑as‑a‑Judge~\cite{zheng2023judging} that assigns coarse scores to free‑form outputs, despite known issues of inconsistency and reward hacking~\cite{jayalath2025compute}. Therefore, injecting reasoning into open‑ended problem generation is challenging.

In this paper, we propose a problem generator that explicitly reasons to plan problem structures before synthesis and dynamically adapts difficulty to the solver's capabilities (see \cref{fig:case_study_lines_full} for a case study). To initialize this, we construct related problem pairs augmented with intermediate problem-design CoT. Although creative \textit{induction} is difficult for current models, they excel at \textit{abduction} given specific contexts~\cite{dai2019bridging}. Unlike standard synthesis ($Q \rightarrow \text{CoT} \rightarrow A$), our cold-start data follows $Q_1 \rightarrow \text{CoT} \rightarrow Q_2$. By using two related subquestions from a multi-part problem, we recover the design rationale connecting them. The later question ($Q_2$) acts as a concrete target, reducing the ambiguity of generating both the rationale and a novel problem from scratch. Specifically, we filter middle- and high-school math datasets for multi-part questions, which naturally provide related pairs that incrementally increase in difficulty around a core concept. A reasoning model then uses a reverse-engineering prompt to extract the \emph{latent CoT} behind these pairs, generating SFT data. This bootstrapping process establishes the reasoning format, enabling the generator to create increasingly difficult problems and preparing it for subsequent RL.

To avoid relying on LLM-as-a-Judge, we use the solver's feedback on synthetic problems as the reward signal. The reward combines the solver's answer consistency on seed and synthesized problems to guide difficulty adjustment relative to the solver's current competence. The generator thus learns to simplify reasoning steps or introduce additional constraints according to the solver's learning needs. Since RLVR benefits from appropriately challenging tasks~\cite{guo2025synthetic}, this feedback aligns problem generation with the objective of improving solver~performance.

Extensive experiments across 10 mathematical and general reasoning benchmarks show that our framework achieves a cumulative average improvement of 3.39 percentage points over seed-only training after three rounds. The framework generalizes across language and vision-language models. With the same cold start, our method outperforms preference-based reward optimization by 2.18 points on mathematical reasoning tasks, highlighting the value of solver-adaptive problem generation. Moreover, a solver trained on synthesized data provides improved feedback for subsequent generator training, enabling co-evolution and further performance gains.

In summary, our main contributions are threefold:
\begin{itemize}
\item We introduce a reasoning-driven problem generator, initialized with problem-design CoT recovered from related subquestions in human-authored multi-part problems.
\item We develop a seed-conditioned reward that uses solver feedback to guide difficulty adjustment, aligning generated problems with the solver's current capabilities.
\item We demonstrate the effectiveness of the framework across mathematical, general-reasoning, and vision-language tasks, with controlled comparisons of data-synthesis methods and analyses of reward components, generation quality, feedback reliability, and sensitivity to the labeling model.
\end{itemize}

\section{Related Work}
\label{sec:related}

\noindent\textbf{Data synthesis in post-training.}
Synthetic data generation, including contrastive or continual multimodal synthesis~\cite{jiao2025img,wu2025synthetic}, text-to-image augmentation for data-efficient multimodal training~\cite{yin2026sparrow}, and inferential or commonsense visual question generation~\cite{bi2025inferential}, expands post-training data when curated annotations are scarce. Unlike visual question generation from images or cross-modal augmentation, we synthesize reasoning problems whose difficulty adapts to an evolving solver. Existing pipelines fall into two broad categories.
(i) Augmentation-based synthesis rewrites seed questions, samples new problems, or generates rationales~\cite{toshniwal2024openmathinstruct,yu2024metamath,yue2024mammoth,zhou2024jiuzhang3,luo2025wizardmath,yu2025cot}. Some pipelines calibrate difficulty using auxiliary models or offline scoring~\cite{pei2025scalediff}. These approaches scale data without typically training explicit problem-design reasoning, often yielding shallow variants or requiring costly candidate generation and selection~\cite{tang2024mathscale,lu2024mathgenie}.
(ii) Iterative self-training and bootstrapping repeatedly generate solutions, select correct or self-corrected trajectories, and fine-tune the solver~\cite{zelikman2022star,zhang2024small,setlur2024rl,wu2025thinking}. Like multimodal reasoning frameworks that apply causal CoT, neural logic, or collaborative inference during answer derivation~\cite{li2026visual,xue2025linin,wen2025enhanced}, these loops optimize solving with verifiable correctness rather than training a generator to pose~problems.

Downstream gains also depend on sample utility and difficulty progression: visual dataset distillation prioritizes crucial samples~\cite{li2025boosting}, while curriculum learning schedules easier-to-harder examples in self-supervised representation learning~\cite{xiao2021explore} and self-paced visual grounding~\cite{xiao2024clipvg}. Rather than selecting or ordering fixed samples, we combine problem-design SFT with solver-feedback RL to generate reasoning-driven problems of adaptive difficulty for solver~training.

\noindent\textbf{Label-free reinforcement learning.}
Large reasoning models, exemplified by DeepSeek‑R1~\cite{guo2025deepseek}, demonstrate that Reinforcement Learning with Verifiable Rewards (RLVR) can markedly improve reasoning capabilities. Unlike Reinforcement Learning from Human Feedback (RLHF), which primarily targets open‑domain instruction alignment~\cite{li2025vl,zhang2025critic,wang2025mv}, RLVR is specifically designed to elicit long CoT for problem solving~\cite{wang2025socratic}. Recent work on label‑free RL in verifiable domains has shown promising results: Test-Time RL~\cite{zuo2025ttrl} uses self-consistency with majority voting to approximate labels for RL fine-tuning in mathematics. Meanwhile, Absolute Zero~\cite{zhao2025absolute} improves language models through self-play on math and coding tasks, progressively solving increasingly complex problems generated by the model itself. Although these strategies are reference-free, they are restricted to domains with verifiable answers. In contrast, R-Zero~\cite{huang2025r} enables LLMs to generate training data entirely from scratch, without any human-curated tasks or labels. However, whereas R-Zero rewards problems near the solver's decision boundary, our reward is conditioned on the solver's accuracy on the seed problem, thereby encouraging directional difficulty adjustment.

In domains lacking verifiable answers, existing approaches often score outputs against references. For example, VeriFree~\cite{zhou2025reinforcing} derives a reward without an external verifier by computing the reference probability from the generated reasoning chain under the initial policy. Similarly, RLMT~\cite{bhaskar2025language} elicits long CoT and applies preference-based online RL. Notably, our approach requires no trained reward model. Furthermore, while most prior work focuses on strengthening solvers~\cite{li2024synthesize}, few studies explore label-free RL to enhance reasoning capabilities in data synthesis generators.

\section{Preliminary}

\noindent\textbf{Group relative policy optimization.}
GRPO~\cite{shao2024deepseekmath} is a memory‑efficient variant of PPO~\cite{schulman2017proximal} that replaces the value network with a group baseline. For each question $q$, we sample $G$ rollouts $o_1, \dots, o_G$ from the current policy $\pi_{\theta_{\text{old}}}$ and optimize
\begin{equation}
\label{eq:grpo}
\begin{aligned}
J(\theta) = \mathbb{E}_{q,\{o_i\}}\Biggl[&\frac{1}{G} \sum_{i=1}^G \frac{1}{|o_i|}
    \sum_{t=1}^{|o_i|} L_t(\theta) \\
    &{}-\beta\,\mathrm{KL}(\pi_\theta\,\|\,\pi_{\text{ref}})\Biggr].
\end{aligned}
\end{equation}
Here $L_t$ denotes the clipped importance-weighted surrogate,
\begin{equation}
\begin{aligned}
L_t(\theta)&=\min\!\Bigl(r_t(\theta)\hat A_i,\\
&\qquad\mathrm{clip}(r_t(\theta),1-\epsilon,1+\epsilon)\hat A_i\Bigr),\\
\hat A_i&=\frac{R(q,o_i)-\bar R_G}{\sigma_G},
\end{aligned}
\end{equation}
where $r_t$ is the token probability ratio between the current and sampling policies, and $\bar R_G$ and $\sigma_G$ denote the group reward mean and standard deviation, respectively. The KL term regularizes the current policy against the initial reference policy. Group-normalized rewards yield advantage estimates without a learned value function.

\noindent\textbf{Reinforcement learning with verifiable rewards.}
The verifiable reward compares the answer extracted from the model’s output with the ground-truth answer. For example, we instruct the model to place its final answer inside \verb|\boxed{·}| and use a regular expression to extract it. Formally, for a model output $\tau$ on question $q$, the reward is
\[
R(\tau) = 
\begin{cases}
1, & \text{if $\tau$ outputs the correct}\\[-2pt]
   & \text{final answer to $q$}, \\
0, & \text{otherwise}.
\end{cases}
\]
This design substantially mitigates reward hacking and enables scalable RL training~\cite{yan2025learning}.

\section{Methodology}

\begin{figure*}[!t]
    \centering
    \includegraphics[width=\textwidth]{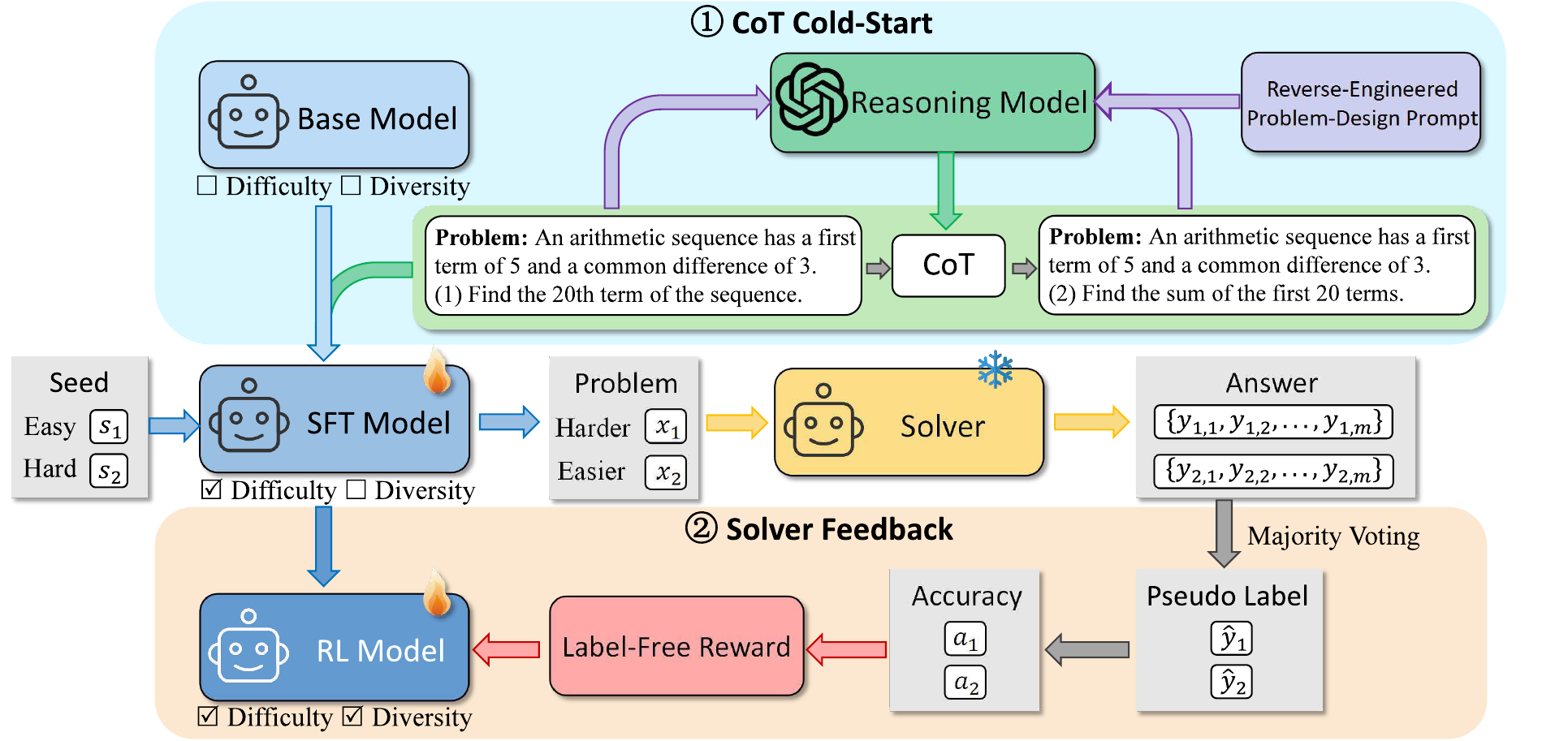}
    \caption{{\bf Training pipeline for our problem generator.} A reasoning model recovers problem-design CoT from paired subquestions to initialize the generator. The generator synthesizes problems from seeds, and the solver answers each problem multiple times. Majority-answer consistency provides the feedback used to compute the seed-conditioned training reward.
    }
    \label{fig:pipeline}
\end{figure*}

Our framework comprises a problem generator, a solver, and a seed set (\cref{fig:pipeline}). We initialize the generator with problem-design CoT recovered from paired subquestions (\cref{sec:sft}), then train it with a label-free reward based on solver answer consistency (\cref{sec:rl}). Finally, an external reasoning model labels the generated problems for downstream solver training (\cref{sec:solver}). The resulting solver provides updated feedback for subsequent rounds of generation, as summarized in \cref{alg:framework}.

\subsection{Cold-Starting from Multi-Part Questions}\label{sec:sft}
Education experts design problems around specific knowledge points and targeted skills. In contrast, model-generated problems often lack well-structured reasoning. RLVR has recently gained traction for eliciting reasoning in base models~\cite{zeng2025simplerl}. However, applying RLVR directly to a problem generator, which is an open-ended task, creates a vast search space and often leads to violations of reasoning and formatting constraints. To provide a stable initialization and reward signal, we first cold-start the policy. This yields formatted reasoning steps and substantially improves sample efficiency.

The challenge is to construct reasoning chains that explain the design of a new problem. Directly prompting a strong model to generate a question provides limited supervision for choosing the underlying transformation. Related problem pairs provide a concrete target for recovering the design rationale through abduction~\cite{dai2019bridging}: inferring the latent design logic that connects two related problems.

Randomly pairing problems to derive an intermediate CoT is unreliable because the pairs may not share a common concept. Instead, we curate authentic middle- and high-school mathematics problems and retain only multi-part questions with several subquestions that progressively build on a single core idea. Because subquestions in a multi-part item are written around a shared stem and usually progress in difficulty, the later subquestion provides a concrete target for recovering the latent design rationale. This makes the cold-start data less arbitrary than prompting a model to invent both the rationale and the follow-up problem. We exclude proofs, routine computations, and very simple items. For each multi-part question, we extract the shared stem and pair it with each subquestion to form problem pairs. Using a structured prompt (Fig.~\ref{fig:template}), we instruct a reasoning model to act as an expert, starting from an example and naturally proposing a follow-up problem. The prompt enforces a three-step workflow: analyze \texttt{Problem1}, choose a transformation strategy, and instantiate it as a follow-up. This yields 34k SFT pairs in the form \texttt{Problem1} $\rightarrow$ (\texttt{CoT}+\texttt{Problem2}). We then apply SFT to initialize the problem generator and teach it to follow the same reasoning format.

\begin{figure}[!t]
\centering

\begin{tcolorbox}[
  colback=blue!5!white,
  colframe=blue!40!black,
  arc=1.5mm,
  auto outer arc,
  boxsep=0.8mm,
  left=1.2mm,right=1.2mm,top=1.0mm,bottom=1.0mm
]
\footnotesize
\raggedright

\setlist[itemize]{leftmargin=1.2em,itemsep=0.15em,topsep=0.15em,parsep=0em,partopsep=0em}
\setlist[enumerate]{leftmargin=*,itemsep=0.25em,topsep=0.2em,parsep=0em,partopsep=0em}

You are a {\bf senior mathematics problem creation expert}. Your task is to derive the creative process
from ``Problem 1'' and its ``Solution 1'' to ``Problem 2'', reconstructing the creative thinking chain.

You must pretend that you do not know ``Problem 2'' at the initial thinking stage. Your output needs to
completely and logically demonstrate how an expert would start from ``Problem 1'', through analysis,
conception, and evolution, to finally happen to design ``Problem 2''.

\textbf{Output Format}
\begin{enumerate}
\item \textbf{Analyze the original problem (Problem 1):}
  \begin{itemize}
    \item Core knowledge points: Briefly list the key concepts / theorems / techniques examined in Problem 1.
    \item Solution characteristics: Summarize the solution style and key step types (high-level description).
  \end{itemize}

\item \textbf{Conceive new problem direction:}
  \begin{itemize}
    \item Problem creation strategy: Specify the adopted strategy (deepening core knowledge points / changing conditions / introducing parameters / contextualization / integrating multiple knowledge points).
    \item Conception process: Use highly summarized thinking to explain why this strategy was chosen and the expected examination ability (without expanding reasoning).
  \end{itemize}

\item \textbf{Derive and form new problem (Problem 2):}
  \begin{itemize}
    \item Specific evolution: Summarize the key changes from original condition A to new condition B, and the resulting change in solution path from method X to method Y (using general terms).
  \end{itemize}
\end{enumerate}

\end{tcolorbox}
\caption{Recover the latent problem-construction logic from paired sub-questions, using the earlier one to infer a rationale for designing the follow-up.}
\label{fig:template}
\end{figure}

\subsection{Label-Free RL with Solver Feedback}\label{sec:rl}

The generator aims to synthesize problems that improve solver performance. Measuring these gains directly requires retraining and evaluating the solver for each candidate, which is costly during rollouts. We therefore use problem difficulty as a feedback signal. Appropriately challenging problems are effective for RLVR~\cite{huang2025r,liang2025sws}: groups with both correct and incorrect responses provide relative advantages, whereas uniform rewards yield zero task advantage.

For each seed, the generator receives the solver's consistency score and a corresponding difficulty-adjustment instruction: simplify for scores below 0.3, preserve difficulty between 0.3 and 0.7, and increase difficulty above 0.7. For a generated problem $q$, the solver samples $m$ responses. Let $\mathcal{P}(q)$ denote the index set of responses with extractable final answers, and let $\hat y$ denote the most frequent answer after grouping mathematically equivalent answers. We define
\begin{equation}
\label{eq:consistency}
a(q)=\frac{1}{|\mathcal{P}(q)|}\sum_{j\in\mathcal{P}(q)}\mathbbm{1}[y_j\equiv\hat y].
\end{equation}
We assign a score of zero when no final answer can be extracted and set consistency values at or below 0.1 to zero. This score serves as a label-free proxy for the solver's competence on a problem. We evaluate its relationship with accuracy on MATH and on generated problems throughout RL in \cref{sec:consistency}.

Let $a_{\mathrm{ori}}$ and $a_{\mathrm{new}}$ denote the consistency scores for the seed and generated problems, respectively. We define the reward as
\begin{equation}
\label{eq:r_acc}
\begin{aligned}
R_{\text{acc}} ={}& 1 - \left| a_{\text{new}} - \left( 1 - a_{\text{ori}} \right) \right|\\
    &{}+\min\left( a_{\text{new}},\, 1 - a_{\text{new}} \right),
\end{aligned}
\end{equation}
The first term encourages difficulty inversion by moving $a_{\mathrm{new}}$ toward $1-a_{\mathrm{ori}}$, while the second rewards intermediate consistency and peaks at $a_{\mathrm{new}}=0.5$. The inversion term specifies an adjustment relative to the seed, and the boundary term favors problems near the solver's current competence boundary. Together, they provide a training signal for simplifying reasoning steps or introducing additional constraints according to the solver's ability. We analyze the resulting optimal interval in \cref{sec:reward_interpretation}.

In addition, we incorporate a format reward to enforce output constraints: the model should place its problem-construction rationale first within \texttt{<think>\allowbreak</think>} and then provide the final question within \texttt{<question>\allowbreak</question>}. Responses that fail to produce a valid question are deemed invalid. This constraint standardizes outputs and accelerates convergence. The final generator reward is:
\begin{equation}
\label{eq:reward}
R_{\text{gen}} =
\begin{cases}
-1, & \text{if invalid}, \\[4pt]
0.9 \cdot R_{\text{acc}} + 0.1 \cdot R_{\text{format}}, & \text{if question},
\end{cases}
\end{equation}
where $R_{\text{format}} \in \{0,1\}$ is a bonus indicating compliance with the output constraints.

We train the policy using the GRPO objective (Eq.~\ref{eq:grpo}) and adopt asymmetric clipping ($\epsilon_{\text{high}} > \epsilon_{\text{low}}$). This permits larger policy updates for high-advantage, promising, or novel samples, thereby avoiding premature clipping and improving exploration~efficiency.

\subsection{Problem Solver Training}\label{sec:solver}
We assess the generator by training a solver on the union of a fixed seed set and the generated problems. We use o3-mini as a shared answer-labeling model across seed-conditioned synthesis methods and apply the same filtering procedure to their generated data. After extracting the questions and boxed answers, we discard problems identified as unsolvable and proof-based problems unsuitable for answer verification. The solver is then trained with GRPO and answer-based rewards. Sensitivity to the labeling model is examined in \cref{sec:labeler_sensitivity}; training and synthesis details are provided in \cref{sec:training_details}.

\subsection{Interpreting the Solver-Adaptive Reward}
\label{sec:reward_interpretation}

\noindent\textbf{Seed-conditioned difficulty adjustment.}
The complementary target $1-a_{\mathrm{ori}}$ reflects the seed's consistency around $0.5$ without introducing an additional parameter. Its displacement from the boundary satisfies $(1-a_{\mathrm{ori}})-0.5=-(a_{\mathrm{ori}}-0.5)$. A high-consistency seed therefore induces a low-consistency target, encouraging the generator to increase difficulty; a low-consistency seed induces a high-consistency target, encouraging simplification. Seeds already near the boundary induce correspondingly small adjustments. This conditioning supplies information about the transformation of each seed that a boundary-only objective does not contain.

\noindent\textbf{An interval of optimal consistency values.}
The combined reward balances the reflected target with the competence boundary. Using $\min(a,1-a)=0.5-|a-0.5|$, we obtain
\begin{equation}
\label{eq:reward_bound}
\begin{aligned}
R_{\mathrm{acc}}
  & = 1.5-\left|a_{\mathrm{new}}-(1-a_{\mathrm{ori}})\right|\\
  &\qquad -\left|a_{\mathrm{new}}-0.5\right|\\
  & \leq 1.5-\left|a_{\mathrm{ori}}-0.5\right|.
\end{aligned}
\end{equation}
The inequality follows from the triangle inequality, with equality whenever $a_{\mathrm{new}}$ lies in the closed interval between $1-a_{\mathrm{ori}}$ and $0.5$. Thus, every consistency value in this interval attains the same maximum reward. Outside the interval, the reward decreases linearly with distance from it. The generator can therefore choose among transformations that move the problem toward an appropriate difficulty range, rather than having to attain one exact score.

For example, a seed with $a_{\mathrm{ori}}=0.2$ yields an optimal interval of $[0.5,0.8]$, whereas a seed with $a_{\mathrm{ori}}=0.8$ yields $[0.2,0.5]$. At $a_{\mathrm{ori}}=0.5$, the interval reduces to a single point. The optimal region is consequently anchored at the competence boundary while retaining a seed-dependent direction and range of adjustment. This explains how the same reward accommodates both simplification and increased difficulty within a single generator.

\noindent\textbf{Connection to the solver's learning signal.}
For a labeled training problem $q$, let $p(q)$ be the probability that a solver rollout receives a correct-answer reward. The variance of this binary reward is $p(q)(1-p(q))$. For $G$ independent rollouts, the probability that a group contains both correct and incorrect answers is
\begin{equation}
\label{eq:mixed_group}
\Pr(\text{mixed rewards}\mid q)=1-p(q)^G-(1-p(q))^G.
\end{equation}
For $G>1$, this probability and the binary reward variance are maximized at $p(q)=0.5$. Groups with mixed outcomes provide the reward contrasts used by GRPO, while all-correct or all-incorrect groups provide no within-group task-reward contrast. This relation motivates the boundary term in our generator reward. The inversion term complements it by determining how to modify a particular seed. During generator training, answer consistency supplies the label-free difficulty estimate; the empirical relationship between this estimate and correctness is examined in \cref{sec:consistency}.

\begin{algorithm}[!t]
\caption{Reasoning-Driven and Solver-Adaptive\\ Data Synthesis}
\label{alg:framework}
\begin{algorithmic}[1]
\REQUIRE Base model $\pi_0$; related pairs $\mathcal{P}=\{(q^{(1)},q^{(2)})\}$; seed set $\mathcal{D}_{\mathrm{seed}}$; rounds $T$; responses per problem $m$
\ENSURE Generator $\pi_G$ and solver $\pi_S$
\STATE For each $(q^{(1)},q^{(2)})\in\mathcal{P}$, recover a design trace $\tau(q^{(1)}\!\rightarrow\!q^{(2)})$
\STATE $\pi_G \leftarrow \mathrm{SFT}\!\left(\pi_0,\{\tau\}\right)$; \quad $\pi_S \leftarrow \pi_0$
\FOR{$t=1,\ldots,T$}
  \STATE For each $q\in\mathcal{D}_{\mathrm{seed}}$, draw $m$ responses from $\pi_S$ and compute $a(q)$ by \eqref{eq:consistency}
  \STATE $\pi_G \leftarrow \mathrm{GRPO}\!\left(\pi_G, R_{\mathrm{gen}}\right)$ with the reward in \eqref{eq:reward}
  \STATE $\mathcal{D}_{\mathrm{syn}} \leftarrow \{\pi_G(q): q\in\mathcal{D}_{\mathrm{seed}}\}$
  \STATE Label $\mathcal{D}_{\mathrm{syn}}$ and remove items that are unparseable, unsolvable, or proof-style
  \STATE $\pi_S \leftarrow \mathrm{GRPO}\!\left(\pi_S,\, \mathcal{D}_{\mathrm{seed}}\cup\mathcal{D}_{\mathrm{syn}}\right)$ with verifiable answer rewards
\ENDFOR
\end{algorithmic}
\end{algorithm}

\section{Experiments}
\label{sec:experiments}
We evaluate downstream performance, isolate data and reward effects, inspect generation quality, and test adaptation across training rounds and seed sets.

\subsection{Experimental Setup}
\label{sec:experimental_setup}

\subsubsection{Models}
Following R-Zero~\cite{huang2025r}, we use Qwen3-4B-Base and Qwen3-8B-Base~\cite{yang2025qwen3} as both problem generators and solvers to assess scaling effects within a single architectural family. To test cross-modal generality, we also evaluate the vision–language model Qwen2.5-VL-7B-Instruct~\cite{bai2025qwen2}. For the VLM, the dataset comprises multimodal reasoning problems, including geometric diagrams and function plots. The generator synthesizes new problems conditioned on the same image.

\subsubsection{Benchmarks and evaluation protocol}
To assess whether reasoning gains transfer beyond the training domain, we evaluate on 10 benchmarks spanning mathematics and general-domain reasoning, each probing a distinct facet of problem solving. The mathematics suite comprises AMC, Minerva~\cite{lewkowycz2022solving}, MATH-500~\cite{hendrycks2measuring}, GSM8K~\cite{cobbe2021training}, Olympiad-Bench~\cite{he2024olympiadbench}, AIME-2024, and AIME-2025. For general-domain reasoning, we evaluate on MMLU-Pro~\cite{wang2024mmlu}, SuperGPQA~\cite{du2025supergpqa}, and BBEH~\cite{kazemi2025big}, following each benchmark’s protocol and scoring with exact~match.

Following standard practice for reasoning models, we generate with temperature 0.6 and top-p 0.95. We report pass@1 accuracy, using the mean accuracy over 128 independently sampled responses per problem (avg@128) on AMC, AIME-2024, and AIME-2025. We use strict exact-match evaluation: a solution is correct only if the boxed answer exactly matches the ground-truth. We do not use any LLM as a programmatic judge for semantic verification.

\subsubsection{Compared baselines}\label{sec:baseline}
Few recent studies have explored training problem generators to reason before synthesizing new problems. We compare with the following representative methods:
\textbf{(i) Seed Set}. Train only on the seed set.
\textbf{(ii) Self-Instruct}~\cite{wang2023self}. Given the seed tasks, prompt the problem generator to produce additional problems directly.
\textbf{(iii) CoT-Self-Instruct}~\cite{yu2025cot}. Instruct the problem generator to first reason and plan via CoT on the seed tasks, then generate new synthetic prompts of comparable quality and complexity for training.
\textbf{(iv) R-Zero}~\cite{huang2025r}. Reward the problem generator for proposing problems at the boundary of the solver’s ability (\ie., those yielding 50\% accuracy).
\textbf{(v) RLMT}~\cite{bhaskar2025language}. A recent method that uses preference-based reward models for open-ended tasks. It requires the problem generator to produce extended CoT reasoning before posing problems and optimizes it with online RL against a reward model (\eg, Skywork-v1-Llama-3.1-8B-v0.2~\cite{liu2024skywork}) as in RLHF.

\subsubsection{Training and synthesis details}
\label{sec:training_details}
We train the problem generator with reinforcement learning on the MATH training set~\cite{hendrycks2measuring}, and randomly sample 4,000 problems as a seed set for subsequent synthesis to train the solver. For a fair comparison, all seed-conditioned mathematical synthesis methods use the same 4,000 seed problems, a shared labeling model, and identical filtering and downstream solver-training procedures. We separately evaluate released mathematical data in \cref{sec:math_baselines}. The VLM generator is trained on the MMK12 training set~\cite{meng2025mm}.

\noindent\textbf{Generator training.}
During SFT, we use a learning rate of $1\times10^{-5}$, a batch size of 120, and 1,200 steps. During RL, we use four rollouts, set the KL penalty coefficient to $1\times10^{-3}$, limit the maximum response length to 6,144 tokens, and train for 20 steps. The consistency score is computed from $m=10$ solver responses per problem.

\noindent\textbf{Synthesis and labeling.}
The trained generator augments each seed once using the prompts in Figs.~\ref{fig:Self-Instruct} and~\ref{fig:our_template}, yielding 4,000 candidate problems. We extract the content within the \texttt{<question>} tags as the problem text. Parsing and solvability filtering lead to different numbers of retained problems across methods. Answers are labeled with o3-mini. In the main results, every problem that passes these filters is combined with the original seed set for solver training. The controlled comparison in \cref{sec:fixed_size} uses a different collection rule: sampling continues from the same seeds until each method retains 4,000 synthetic~problems.

\noindent\textbf{Solver training.}
The problem solver is trained with GRPO using rule-based, verifiable rewards. The solver must place its final answer in the \verb|\boxed| format. We use AdamW with a cosine learning rate schedule and a peak learning rate of $1\times10^{-6}$, and train for 100 steps with a batch size of 128. We generate eight rollouts per prompt with temperature 1.0, top-p 0.99, and a maximum sequence length of 4,096 tokens. The KL penalty coefficient is $1\times10^{-3}$, and gradients are clipped at a maximum norm of 1.0. We use Fully Sharded Data Parallel (FSDP) with full parameter sharding and optional CPU offloading for parameters and optimizer states. GRPO experiments use the EasyR1 codebase~\cite{zheng2025easyr1}, and MathRuler~\cite{mathruler} serves as the program verifier. The solver prompt is provided in Fig.~\ref{fig:solver}. Prompt templates and the corresponding training settings are also collected in Appendix~\ref{app:detail}.

\subsection{Main Results}\label{sec:results}
We first compare solver performance across mathematical, general-domain, and multimodal reasoning tasks.

\subsubsection{Mathematical reasoning}
\begin{table*}[!t]
\centering
\caption{Performance Comparison on Mathematical Reasoning Benchmarks. Pass@1 accuracy (\%). AMC, AIME24, and AIME25 use the mean of 128 samples per problem (avg@128). Indented rows are variants of the method above (Sec.~\ref{sec:baseline}). Bold and underlined values are the best and second-best scores.}
\label{tab:math_results}
\resizebox{0.80\textwidth}{!}{%
\footnotesize
\setlength{\tabcolsep}{3.5pt}
\renewcommand{\arraystretch}{1.12}
\begin{tabular}{
    l
    *{8}{S[table-format=2.2]}
}
\toprule
 & {\textbf{MATH}} & {\textbf{GSM8K}} & {\textbf{AMC}} & {\textbf{Minerva}} & {\textbf{Olympiad}} & {\textbf{AIME24}} & {\textbf{AIME25}} & {\textbf{Avg.}} \\
\midrule
\rowcolor{gray!10}
\multicolumn{9}{c}{\textit{\textbf{Qwen3-4B-Base}}} \\
Base Model & 66.40 & 70.49 & 44.68 & 17.71 & 31.01 & 9.38 & 7.22 & {35.27} \\
Seed Set & 79.60 & 92.79 & 56.35 & 31.73 & 39.76 & 14.52 & 12.02 & 46.68 \\
\midrule
Self-Instruct~\cite{wang2023self} & 77.40 & 92.79 & 55.49 & $\boldsymbol{32.84}$ & 37.69 & 12.17 & 7.40 & {45.11} \\
\quad + CoT-Self-Instruct~\cite{yu2025cot} & 78.20 & $\boldsymbol{93.47}$ & 56.00 & {$\underline{32.47}$} & 38.58 & 11.94 & 7.04 & {45.39} \\
\quad + RLMT~\cite{bhaskar2025language} & 79.60 & 91.05 & 55.80 & 31.25 & 39.26 & 13.62 & 10.36 & 45.85 \\
\quad + R-Zero~\cite{huang2025r} & 80.80 & 92.79 & 60.26 & 30.63 & $\boldsymbol{44.81}$ & 17.23 & 14.55 & 48.72 \\
\quad + Solver Feedback (Ours) & $\underline{81.20}$ & 92.41 & $\boldsymbol{60.93}$ & 30.63 & 43.47 & $\boldsymbol{18.14}$ & $\boldsymbol{15.30}$ & {$\underline{48.87}$} \\
\midrule
CoT Cold-Start (Ours) & 77.80 & 91.96 & 57.70 & 31.73 & 41.25 & 14.62 & 10.58 & {46.52} \\
\quad + RLMT~\cite{bhaskar2025language} & 80.80 & 91.81 & 57.64 & 30.26 & 40.50 & 15.43 & 13.01 & 47.06 \\
\quad + R-Zero~\cite{huang2025r} & 78.40 & 92.79 & 55.73 & 30.26 & 39.91 & 11.57 & 7.53 & {45.17} \\
\quad + Solver Feedback (Ours) & $\boldsymbol{82.00}$ & $\underline{92.87}$ & $\underline{60.52}$ & $\boldsymbol{32.84}$ & $\underline{44.36}$ & $\underline{17.44}$ & $\underline{14.65}$ & {$\boldsymbol{49.24}$} \\
\midrule
\rowcolor{gray!10}
\multicolumn{9}{c}{\textit{\textbf{Qwen3-8B-Base}}} \\
Base Model & 67.80 & 80.35 & 46.03 & 16.97 & 34.42 & 11.24 & 10.45 & {38.18} \\
Seed Set & 83.40 & 93.10 & 64.70 & 32.84 & 44.51 & 21.27 & 16.71 & 50.93 \\
\midrule
Self-Instruct~\cite{wang2023self} & 83.40 & 93.02 & 64.52 & $\underline{35.79}$ & 45.40 & $\underline{22.00}$ & $\underline{17.65}$ & $\underline{51.68}$ \\
\quad + CoT-Self-Instruct~\cite{yu2025cot} & 80.20 & 93.40 & 58.99 & $\boldsymbol{37.27}$ & 39.61 & 14.57 & 14.65 & 48.38 \\
\quad + RLMT~\cite{bhaskar2025language} & 82.60 & 92.87 & 63.95 & 34.56 & 44.15 & 20.94 & 16.80 & 50.84 \\
\quad + R-Zero~\cite{huang2025r} & 83.40 & $\underline{93.63}$ & 61.00 & 33.21 & 44.07 & 17.78 & 15.51 & 49.80 \\
\quad + Solver Feedback (Ours) & 82.80 & 93.40 & 64.78 & 29.89 & $\underline{46.59}$ & 21.51 & 17.00 & 50.85 \\
\midrule
CoT Cold-Start (Ours) & 81.80 & 93.02 & 62.22 & 33.21 & 41.39 & 18.61 & 15.56 & 49.40 \\
\quad + RLMT~\cite{bhaskar2025language} & 83.40 & 93.25 & $\underline{64.97}$ & 32.47 & 44.21 & 21.95 & 17.44 & {51.10} \\
\quad + R-Zero~\cite{huang2025r} & $\boldsymbol{86.20}$ & $\boldsymbol{93.70}$ & 62.84 & 33.21 & $\boldsymbol{46.88}$ & 21.12 & 16.63 & 51.51 \\
\quad + Solver Feedback (Ours) & $\underline{83.80}$ & 93.32 & $\boldsymbol{67.42}$ & $\underline{35.79}$ & $\underline{46.59}$ & $\boldsymbol{22.08}$ & $\boldsymbol{18.17}$ & $\boldsymbol{52.45}$ \\
\bottomrule
\end{tabular}%
}
\end{table*}
As shown in Tab.~\ref{tab:math_results}, our method achieves the highest average accuracy across seven mathematical reasoning benchmarks at both model scales, reaching 49.24\% on the 4B model and 52.45\% on the 8B model. These results exceed seed-only training by 2.56 and 1.52 percentage points, respectively. We summarize the main observations as follows.
(i) \emph{SFT and RL are complementary.} CoT Cold-Start alone is not always the best: it outperforms Self-Instruct on Qwen3-4B but underperforms it on Qwen3-8B. When combined with Solver Feedback, however, the full method consistently yields the best average on both models, suggesting that the reasoning-format prior and the solver-grounded reward work synergistically rather than as independent contributors.
(ii) \emph{Solver feedback versus preference-based rewards.} With the same CoT cold start and Qwen3-4B solver, our method improves the average mathematical accuracy from RLMT's 47.06\% to 49.24\%, a gain of 2.18 percentage points. The reward is computed directly from the target solver's responses, without a separate preference model.

\subsubsection{General-domain reasoning}
\begin{table}[!t]
\centering
\footnotesize
\setlength{\tabcolsep}{3pt}
\renewcommand{\arraystretch}{1.10}
\caption{Performance Comparison on General-Domain Reasoning Benchmarks}
\label{tab:general_reasoning}
\begin{tabular*}{\linewidth}{@{}l@{\extracolsep{\fill}} S S S S@{}}
\toprule
 & {\shortstack{\textbf{Super}\\\textbf{GPQA}}} & {\textbf{BBEH}} & {\shortstack{\textbf{MMLU-}\\\textbf{Pro}}} & {\textbf{Avg.}} \\
\midrule
\rowcolor{gray!10}
\multicolumn{5}{c}{\textit{\textbf{Qwen3-4B-Base}}} \\
Base Model & 20.88 & 7.57 & 37.38 & 21.94 \\
Seed Set & 29.74 & 11.22 & 52.88 & 31.28 \\
\midrule
Self-Instruct & 30.07 & 11.13 & 56.79 & 32.66 \\
\quad + CoT-Self-Instruct & 30.35 & 11.50 & 57.86 & 33.24 \\
\quad + R-Zero & 31.27 & 11.19 & 58.79 & 33.75 \\
\quad + Solver Feedback (Ours)& 31.29 & 11.68 & 59.78 & 34.25 \\
\midrule
CoT Cold-Start (Ours)& 30.65 & $\boldsymbol{11.86}$ & 57.12 & 33.21 \\
\quad + RLMT & 29.70 & 10.75 & 53.28 & 31.24 \\
\quad + R-Zero & 29.15 & 9.93 & 53.67 & 30.92 \\
\quad + Solver Feedback (Ours) & $\boldsymbol{31.33}$ & 11.35 & $\boldsymbol{60.11}$ & $\boldsymbol{34.26}$ \\
\bottomrule
\end{tabular*}
\end{table}
As shown in Tab.~\ref{tab:general_reasoning}, our math-centric approach increases the average general-domain reasoning score on SuperGPQA, BBEH, and MMLU-Pro by 2.98\%, outperforming all other methods. Notably, on MMLU-Pro, our method reaches 60.11\%, surpassing R-Zero by 1.32\%. These gains mirror the patterns observed on mathematical benchmarks, underscoring the effectiveness of the proposed modules. Despite being trained solely on mathematical data, the solver generalizes well, improving performance on tasks that require factual and multi-domain knowledge, as well as complex, cross-disciplinary reasoning.

\subsubsection{Multimodal reasoning}
\begin{table}[!t]
\centering
\footnotesize
\setlength{\tabcolsep}{3pt}
\renewcommand{\arraystretch}{1.10}
\caption{Performance Comparison on Multimodal Reasoning Benchmarks}
\label{tab:vlm}
\begin{tabular*}{\linewidth}{@{}l@{\extracolsep{\fill}} S S S S@{}}
\toprule
 & {\shortstack{\textbf{Math}\\\textbf{Vista}}} & {\shortstack{\textbf{Math}\\\textbf{Vision}}} & {\shortstack{\textbf{Math}\\\textbf{Verse}}} & {\textbf{Avg.}} \\
\midrule
\rowcolor{gray!10}
\multicolumn{5}{c}{\textit{\textbf{Qwen2.5-VL-7B-Instruct}}} \\
Base Model & 68.40 & 23.35 & 43.53 & 45.09 \\
Seed Set & 69.00 & 29.28 & 45.23 & 47.84 \\
\midrule
Self-Instruct~\cite{wang2023self} & 69.80 & 29.60 & 46.40 & 48.60 \\
\quad + CoT-Self-Instruct~\cite{yu2025cot} & 69.30 & 27.63 & 45.58 & 47.50 \\
\quad + R-Zero~\cite{huang2025r} & 69.30 & 26.64 & 44.72 & 46.89 \\
\quad + Solver Feedback (Ours) & $\boldsymbol{70.30}$ & 27.63 & 46.17 & 48.03 \\
\midrule
CoT Cold-Start (Ours) & 69.10 & 28.29 & 45.41 & 47.60 \\
\quad + Solver Feedback (Ours) & 70.10 & $\boldsymbol{30.26}$ & $\boldsymbol{46.65}$ & $\boldsymbol{49.00}$ \\
\bottomrule
\end{tabular*}
\end{table}
To assess generalization, we use the vision-language model Qwen2.5-VL-7B-Instruct as both the problem generator and the solver. Because many problems include images (\eg, geometry and function problems), synthesizing image-conditioned problems is essential. We evaluate the solver accuracy on MathVista~\cite{lu2024mathvista}, MathVision~\cite{wang2024measuring}, and MathVerse~\cite{zhang2024mathverse}. These benchmarks comprehensively assess multimodal mathematical reasoning across diverse visual contexts.
As shown in Tab.~\ref{tab:vlm}, while our individual modules (Solver Feedback and CoT Cold-Start) effectively improve upon the base model, their combination (CoT + Solver Feedback) delivers the largest average gain (+3.91\% over the base model) and achieves the best scores on MathVision and MathVerse. These results confirm that our framework successfully extends beyond text-only domains.

\subsection{Controlled Comparisons of Synthetic Data}
\label{sec:controlled_comparisons}

\subsubsection{Comparison at a fixed training-set size}
\label{sec:fixed_size}
Parsing and solvability filtering leave a different number of usable problems after one candidate per seed. For this comparison, each method continues sampling from the shared 4,000 seeds under the same prompts and filters until at least 4,000 problems pass. We then keep 4,000 of those retained problems and add them to the seeds. All methods follow the same downstream solver-training and evaluation procedures. This setting measures the contribution of the synthesized problems when the number of training examples is fixed.

\Cref{tab:8000_results} reports the full results on seven mathematical benchmarks. Our method achieves the highest average accuracy of 49.90\%. With the same CoT cold start, replacing the boundary-only R-Zero objective with solver feedback improves the average from 47.45\% to 49.90\%. The full method attains the highest scores on MATH, AMC, Olympiad, and AIME25, and ties for the highest score on AIME24. The advantage observed in the main comparison is thus retained when every method contributes the same number of synthetic training problems.

\begin{table*}[!t]
\centering
\caption{Controlled Comparison with a Fixed Training-Set Size. Pass@1 accuracy (\%), with avg@128 on AMC, AIME24, and AIME25. Each method uses the same 4,000 seeds plus 4,000 retained synthetic problems. Indented rows are variants of the method above (Sec.~\ref{sec:baseline}). Bold and underlined values are the best and second-best scores.}
\label{tab:8000_results}
\resizebox{0.80\textwidth}{!}{%
\footnotesize
\setlength{\tabcolsep}{3.5pt}
\renewcommand{\arraystretch}{1.12}
\begin{tabular}{
    l
    *{8}{S[table-format=2.2]}
}
\toprule
 & {\textbf{MATH}} & {\textbf{GSM8K}} & {\textbf{AMC}} & {\textbf{Minerva}} & {\textbf{Olympiad}} & {\textbf{AIME24}} & {\textbf{AIME25}} & {\textbf{Avg.}} \\
\midrule
\rowcolor{gray!10}
\multicolumn{9}{c}{\textit{\textbf{Qwen3-4B-Base}}} \\
Base Model & 66.40 & 70.49 & 44.68 & 17.71 & 31.01 & 9.38 & 7.22 & {35.27} \\
Seed Set & 79.60 & 92.79 & 56.35 & 31.73 & 39.76 & 14.52 & 12.02 & 46.68 \\
\midrule
Self-Instruct~\cite{wang2023self} & 79.60 & 92.57 & 55.16 & 31.99 & 38.07 & 14.53 & 13.96 & 46.55 \\
\quad + CoT-Self-Instruct~\cite{yu2025cot} & 78.60 & $\boldsymbol{93.63}$ & 55.53 & $\boldsymbol{33.82}$ & 37.04 & 10.78 & 11.04 & 45.78 \\
\quad + R-Zero~\cite{huang2025r} & 80.80 & 92.80 & 60.00 & 30.51 & $\underline{44.44}$ & $\underline{17.29}$ & 14.79 & 48.66 \\
\quad + Solver Feedback (Ours) & $\underline{81.40}$ & 92.65 & $\underline{61.09}$ & 30.51 & 43.85 & $\boldsymbol{18.20}$ & $\underline{15.21}$ & $\underline{48.99}$ \\
\midrule
CoT Cold-Start (Ours) & 78.40 & 92.12 & 58.01 & 31.62 & 41.63 & 14.79 & 11.20 & 46.82 \\
\quad + RLMT~\cite{bhaskar2025language} & 81.20 & 92.19 & 58.20 & 29.78 & 41.04 & 15.99 & 13.75 & 47.45 \\
\quad + R-Zero~\cite{huang2025r} & 80.20 & 92.57 & 56.41 & 30.51 & 40.89 & 16.51 & 15.08 & 47.45 \\
\quad + Solver Feedback (Ours) & $\boldsymbol{82.20}$ & $\underline{93.03}$ & $\boldsymbol{61.68}$ & $\underline{32.72}$ & $\boldsymbol{44.89}$ & $\boldsymbol{18.20}$ & $\boldsymbol{16.59}$ & $\boldsymbol{49.90}$ \\
\bottomrule
\end{tabular}%
}
\end{table*}

\subsubsection{Comparison with mathematical data synthesis}
\label{sec:math_baselines}
We additionally compare with MetaMath~\cite{yu2024metamath} and ScaleDiff~\cite{pei2025scalediff} using the same Qwen3-4B solver and an 8,000-example training set consisting of 4,000 seed problems and 4,000 retained synthetic problems. MetaMath uses its official question-only rewriting branch on the same MATH seeds, followed by the shared labeling, filtering, and solver-training procedure. For ScaleDiff, we use the public ScaleDiff-Math generated subset with its released solutions; this comparison evaluates the utility of the released data rather than rerunning DiffGen under a matched generation budget. Both comparisons use the same seven-benchmark evaluation~suite.

\begin{table}[!t]
\centering
\small
\setlength{\tabcolsep}{4pt}
\renewcommand{\arraystretch}{1.18}
\caption{Comparison with Mathematical Data Synthesis. Macro-average accuracy over seven mathematical benchmarks with the same solver and training-set size.}
\label{tab:math_specialized}
\begin{tabular*}{\linewidth}{@{}l@{\extracolsep{\fill}}lr@{}}
\toprule
Method & Comparison setting & Avg. (\%)\\
\midrule
MetaMath & Seed-conditioned rewriting & 46.80\\
ScaleDiff & Released-data utility & 47.30\\
Ours & Seed-conditioned generation & \textbf{49.90}\\
\bottomrule
\end{tabular*}
\end{table}

As shown in \cref{tab:math_specialized}, our method exceeds MetaMath by 3.10 percentage points and ScaleDiff by 2.60 points. MetaMath expands the seed set through question rewriting, while ScaleDiff emphasizes a globally difficult distribution. Our generator instead conditions the difficulty adjustment on the target solver's performance on each seed. The results support the value of solver-adaptive generation for downstream training, complementing the comparisons with preference rewards and boundary-only rewards.

\subsection{Reward and Solver-Feedback Analysis}
\label{sec:reward_analysis}
We examine reward components, solver-consistency reliability, and generation behavior during RL.

\subsubsection{Reward ablation}
\begin{table}[!t]
    \setlength{\tabcolsep}{12pt}
    \centering
    \small
    \renewcommand{\arraystretch}{1.05}
    \caption{Reward Ablation on Seven Mathematical Benchmarks}
    \label{tab:reward_ablation}
    \begin{tabular}{l S[table-format=2.2]}
    \toprule
    \textbf{Reward variant} & \textbf{Avg.} \\
    \midrule
    Boundary only (R-Zero)        & {45.17} \\
    Inversion only                & 47.61 \\
    Full reward (Ours)            & {\textbf{49.24}} \\
    \bottomrule
    \end{tabular}
\end{table}
Our reward function in \cref{eq:r_acc} consists of two terms: an inversion term that promotes difficulty inversion, and a boundary term that targets decision-boundary difficulty. We ablate these components by training the generator with (1) inversion only, (2) boundary only (corresponds to R-Zero), and (3) the full reward. As shown in \cref{tab:reward_ablation}, the inversion-only setting underperforms the full reward. This indicates that the boundary term complements inversion by generating problems of varying difficulty near the solver’s decision boundary, which are valuable for training. The gap between boundary-only and the full reward further indicates that conditioning on the seed problem provides useful information beyond simply targeting the solver's decision boundary.

\subsubsection{Reliability of solver consistency}
\label{sec:consistency}
Since ground-truth answers are unavailable for generated problems during RL rollouts, we use the solver's answer consistency, defined in \cref{eq:consistency}, as an estimate of accuracy. We evaluate the reliability of this proxy on MATH and on synthetic problems generated during RL.

\begin{figure}[!t]
\centering
\includegraphics[width=0.86\linewidth]{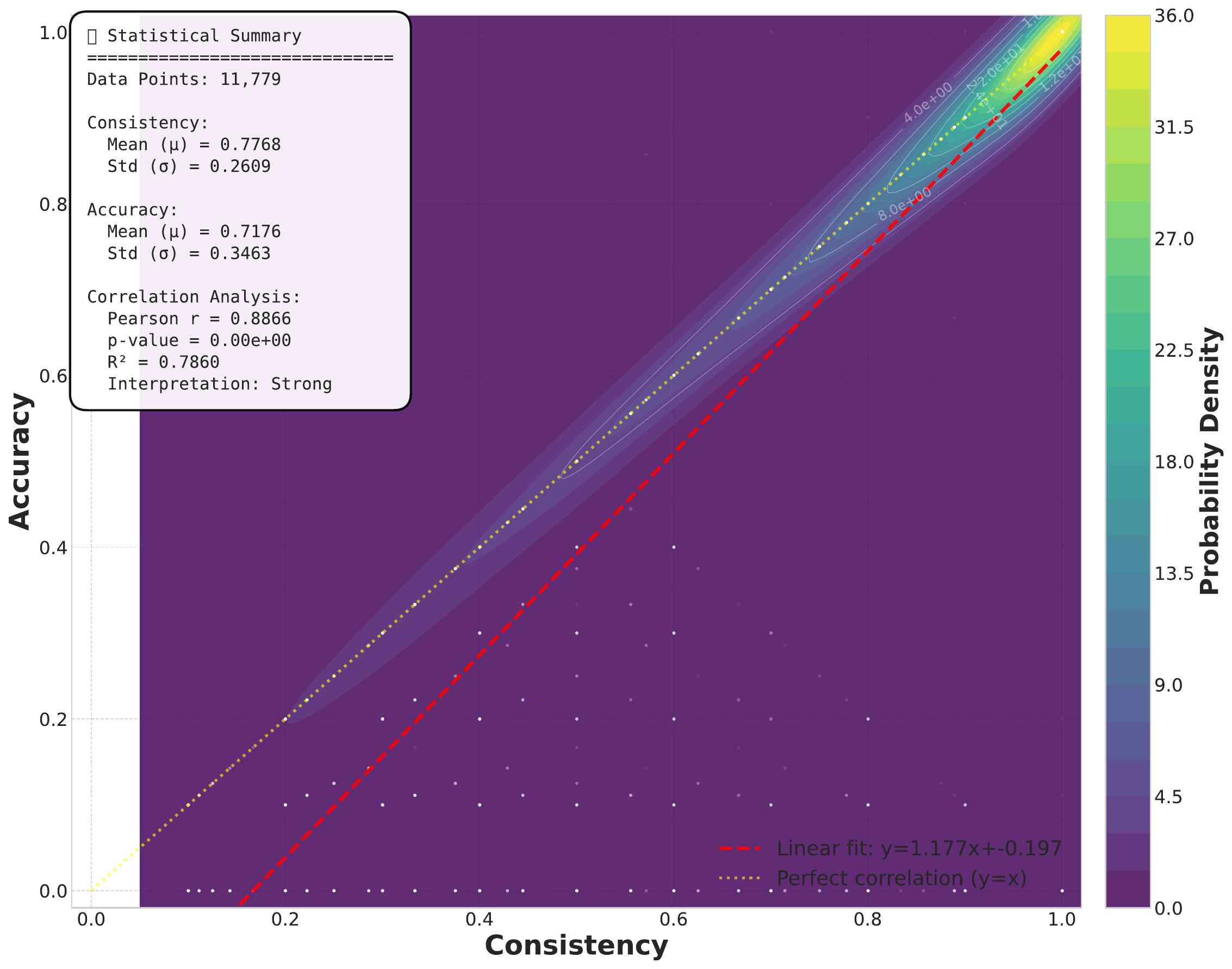}
\caption{\textbf{Contour density plot of solver consistency and accuracy on MATH.} The dashed red line shows the best-fit linear regression, and the yellow dotted line denotes the identity reference ($y=x$). Pearson $r=0.887$ indicates a strong positive relationship.}
\label{fig:consistency}
\end{figure}

Fig.~\ref{fig:consistency} presents a contour density plot illustrating the relationship between consistency and accuracy on MATH. The color map represents the probability density estimated via Gaussian KDE~\cite{silverman2018density}, with lighter regions indicating a higher concentration of data points. The dashed red line shows the best-fit linear regression, $y=1.177x-0.197$, while the yellow dotted line denotes the identity reference ($y=x$). Statistical analysis yields a Pearson correlation coefficient of $r=0.887$ ($p<0.001$) and a Spearman rank correlation coefficient of $\rho=0.954$, indicating a strong positive relationship between consistency and accuracy. The confidently wrong rate (consistency at least 0.8 with an incorrect majority answer) is 1.3\%. These results support the use of solver consistency as an accuracy proxy for generator~training.

We further evaluate consistency on synthetic problems sampled from generator checkpoints at steps 0, 5, 10, and 20. An independent LLM verifier provides reference answers and validity judgments. For each problem, we compute consistency from $m=10$ solver responses and measure verifier-based accuracy as the fraction of extracted answers matching the reference. \Cref{tab:synthetic_consistency} reports the correlations between these two measures, together with the rates of confidently wrong answers, invalid or unsolvable problems, and parseable solver~responses.

\begin{table}[!t]
\centering
\footnotesize
\setlength{\tabcolsep}{2pt}
\renewcommand{\arraystretch}{1.15}
\caption{Reliability of Solver Consistency during Generator RL. Confidently wrong denotes consistency $\geq0.8$ with a majority answer that does not match the verifier's reference answer. Parseable answers are responses with an extractable final answer.}
\label{tab:synthetic_consistency}
\begin{tabular}{lrrrrr}
\toprule
Checkpoint & \shortstack{Pearson\\$r$} & \shortstack{Spearman\\$\rho$} & \shortstack{Confidently\\wrong (\%)} & \shortstack{Invalid /\\unsolvable\\(\%)} & \shortstack{Parseable\\answers\\(\%)} \\
\midrule
Step 0 & 0.78 & 0.81 & 2.1 & 6.4 & 86 \\
Step 5 & 0.74 & 0.79 & 2.6 & 7.1 & 85 \\
Step 10 & 0.76 & 0.80 & 2.4 & 6.8 & 84 \\
Step 20 & 0.73 & 0.77 & 2.9 & 7.5 & 83 \\
\bottomrule
\end{tabular}
\end{table}

Across the evaluated checkpoints, Pearson's $r$ ranges from 0.73 to 0.78, and the confidently wrong rate ranges from 2.1\% to 2.9\%. Although lower than on MATH, the correlation with verifier-based accuracy remains positive throughout generator training, supporting the use of consistency as a difficulty signal for the seed-conditioned reward.

\subsubsection{Training dynamics}
\begin{figure*}[!t]
    \centering
    \includegraphics[width=\textwidth]{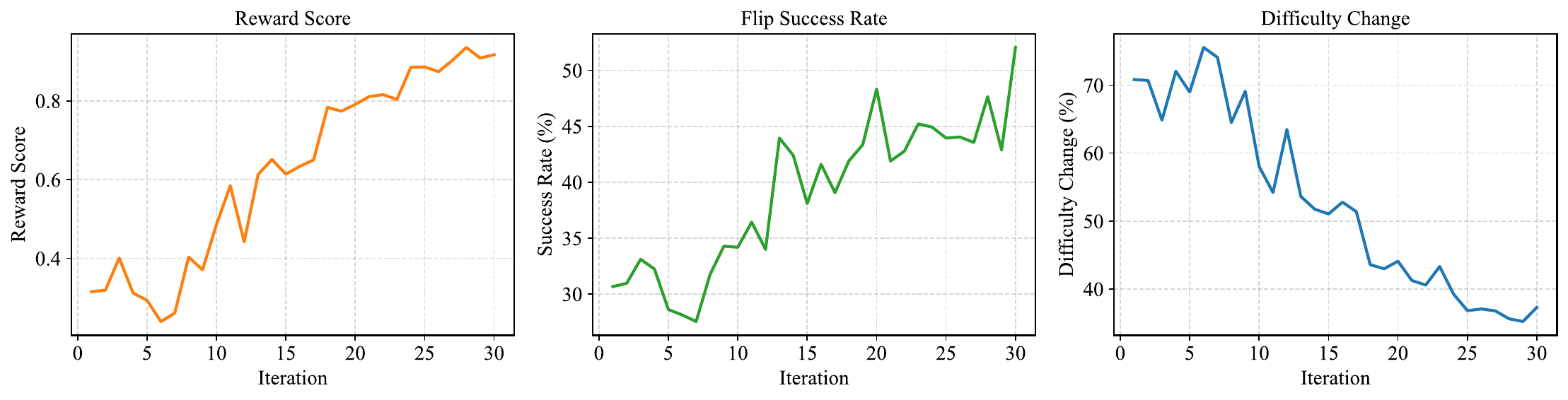}
    \caption{{\bf Training dynamics across iterations}: the problem generator’s reward score, flip success rates for the original and new problems, and changes in difficulty for the original and new problems.}
    \label{fig:reward_three_curves}
\end{figure*}

We plot the training dynamics of the problem generator during RL. As shown in Fig.~\ref{fig:reward_three_curves}, the reward score defined in Eq.~\ref{eq:reward} rises steadily, indicating successful optimization under GRPO. We also track two metrics: (i) flip success rate, the proportion of samples whose difficulty is flipped, meaning the solver’s accuracy crosses the 0.5 threshold from $<$ 0.5 to $\ge$ 0.5 or the reverse; and (ii) difficulty change, the absolute difference between solver accuracy on the original and revised items. The flip success rate increases over training, showing that the generator learns to both harden and simplify problems rather than relying on a single strategy. In contrast, difficulty change decreases, indicating that large early adjustments give way to smaller edits that keep revised items near the 0.5 decision boundary while still inducing flips. Together, these trends confirm effective training and explain why the method works.

\raggedbottom
\subsection{Generation Quality and Case Analysis}
\label{sec:quality_analysis}
\makeatletter
\newenvironment{casecomparison}[2]{%
  \par
  \begingroup
  \normalfont\normalsize
  \def\casecomparisontext{#1}%
  \def\casecomparisonlabel{#2}%
    \def\@captype{figure}%
  \begin{tcolorbox}[enhanced,colback=gray!7,colframe=gray!50,
    boxrule=0.5pt,width=\linewidth,
    coltext=black,
    before skip=6pt,after={\par\nobreak\vspace{3pt}}]
  \small
  \setlength{\parindent}{0pt}
  \setlength{\parskip}{2pt}
  \setlength{\abovedisplayskip}{4pt}
  \setlength{\belowdisplayskip}{4pt}
}{%
  \end{tcolorbox}
  \nopagebreak
  \expandafter\caption\expandafter{\casecomparisontext}%
  \expandafter\label\expandafter{\casecomparisonlabel}\par
  \endgroup
}
\makeatother

\begin{figure*}[!t]
\begin{minipage}{\textwidth}
\begin{casecomparison}{\textbf{Case study under high solver feedback: increased difficulty.} Responses to the line-intersection problem.}{fig:case_study_lines_full}
\textbf{Original problem:} Line $m$ has the equation $y=3x+5$. Line $n$ has the equation $y=kx-7$. Lines $m$ and $n$ intersect at the point $(-4,-7)$. What is the value of $k$?\quad(\emph{Solver feedback: 1.0.})

\textbf{Response of CoT-Self-Instruct:}
\begin{enumerate}[leftmargin=*,itemsep=1pt,topsep=2pt,parsep=0pt]
\item Student's current accuracy rate is 1.0, which is very high.
\item \textbf{The original problem was:} Find the value of k.
\item \textbf{Difficulty level:} high (since 1.0 \textgreater{} 0.7)
\item \textbf{Rule to apply:} Increase difficulty - add complexity
\end{enumerate}
\par
$[\ldots]$\par
Now, let's introduce an additional constraint:\par
\par
\textbf{Generated problem:} Line \( m \) has the equation \( y = 3x + 5 \). Line \( n \) has the equation \( y = kx - 7 \). Line \( p \) has the equation \( y = mx + b \), where \( m \) and \( b \) are constants. Lines \( m \) and \( n \) intersect at the point \( (-4, -7) \). Lines \( m \), \( n \), and \( p \) intersect at the same point. What is the value of \( k \)?\par

\par\noindent\rule{\linewidth}{0.1pt}

\textbf{Response of our final model (condensed):}
\begin{enumerate}[leftmargin=*,itemsep=1pt,topsep=2pt,parsep=0pt]
\item \textbf{Analyze the student's performance:} The student's accuracy is 1.0, and the seed tests direct substitution into a line equation. A more challenging problem should introduce an additional algebraic step.
\item \textbf{Choose the adjustment:} Use coupled equations rather than a single equation that only requires rearrangement. Let one equation provide an expression for substitution into the other, adding a reasoning step within the same topic.
\item \textbf{Formulate the new problem:} Use $y=2x-5$ together with $2y+3x=12$. The first equation isolates $y$, while the second requires recognizing and applying the substitution to determine their intersection.
\end{enumerate}
\textbf{Generated problem:} A line has equation $y = 2x - 5$. Another line has equation $2y + 3x = 12$. Find the value of the $x$-coordinate of their point of intersection.\par

\end{casecomparison}

In Fig.~\ref{fig:case_study_lines_full}, CoT-Self-Instruct adds a third line but retains the known intersection point, so the requested parameter is still obtained from $-7=-4k-7$. Our final model considers alternative structures before selecting two coupled equations that require substitution. Solving $2(2x-5)+3x=12$ gives $x=22/7$. The revision replaces a supplied intersection with one that must be derived, adding an algebraic reasoning step within the same topic.
\end{minipage}
\end{figure*}

We inspect problem quality and labeling sensitivity, then illustrate structural adaptation with a concrete case.
\subsubsection{Quality of synthesized problems}
\label{sec:generation_quality}

\noindent\textbf{Audit protocol.}
To examine generation quality directly, we sample 100 candidate problems per method from the same seed pool for Ours, R-Zero, and CoT-Self-Instruct. Two human annotators independently assess each candidate and cross-check their annotations using a fixed rubric. The audit evaluates the generated problems themselves, complementing the downstream solver results.

We assess six properties. A \emph{well-posed} problem is self-contained and coherent, without missing context or figures. A \emph{solvable} problem admits a reasonable finite answer and is neither under-specified nor clearly unsolvable. \emph{Ambiguity} covers unclear statements or multiple equally valid answers. A \emph{seed near-duplicate} differs only in wording or numerical values, or is nearly semantically equivalent to its seed. A \emph{strategy change} requires a material change in the knowledge or solution steps needed to solve the seed. Finally, \emph{label--verifier agreement} measures whether the externally generated answer agrees with an independent verifier.

\begin{table}[ht]
\centering
\footnotesize
\setlength{\tabcolsep}{3pt}
\renewcommand{\arraystretch}{1.1}
\caption{Quality Audit of Generated Problems. All values are percentages. Each method contributes 100 candidate problems sampled from the same seed pool.}
\label{tab:generation_quality}
\resizebox{\columnwidth}{!}{%
\begin{tabular}{lrrrrrr}
\toprule
Method & \shortstack{Well-posed\\$\uparrow$} & \shortstack{Solvable\\$\uparrow$} & \shortstack{Ambiguous\\$\downarrow$} & \shortstack{Seed near-\\duplicate $\downarrow$} & \shortstack{Strategy\\changed $\uparrow$} & \shortstack{Label--verifier\\agreement $\uparrow$}\\
\midrule
CoT-Self-Instruct & 82 & 80 & 8 & 21 & 19 & 83\\
R-Zero & 79 & 76 & 9 & 18 & 22 & 81\\
Ours & \textbf{86} & \textbf{84} & \textbf{5} & \textbf{4} & \textbf{31} & \textbf{88}\\
\bottomrule
\end{tabular}%
}
\end{table}

\noindent\textbf{Validity and structural diversity.}
As shown in \cref{tab:generation_quality}, our generator produces a higher proportion of well-posed and solvable problems, reaching 86\% and 84\%, respectively. Its ambiguity rate is 5\%, and label--verifier agreement reaches 88\%. The seed near-duplication rate falls to 4\%, compared with 18\% for R-Zero and 21\% for CoT-Self-Instruct, while the strategy-change rate increases to 31\%. These results connect the downstream gains to improved problem validity and more substantial transformations of the seeds.

\noindent\textbf{Solver-conditioned adaptation.}
We also evaluate whether each generated problem changes difficulty in the intended direction. For seeds with consistency below 0.3, a successful adjustment increases consistency; for seeds above 0.7, it decreases consistency. For intermediate seeds, consistency must remain within 0.2 of its original value. Across generator checkpoints at steps 0, 5, 10, and 20, this directional success rate averages 0.61, with values ranging from 0.54 to 0.69. On the single pass used for the main results, our generator retains 68\% of candidates after parsing and solvability filtering. The fixed-size comparison in \cref{sec:fixed_size} therefore draws further candidates from the same seeds until 4,000 problems are retained. These diagnostics complement the flip success rate in Fig.~\ref{fig:reward_three_curves}: they evaluate the simplify/preserve/harden instruction associated with each seed, while flip success records movement across the central difficulty threshold.

The case in \cref{sec:case_study} illustrates increased difficulty through a new dependency between solution steps. Appendix~\ref{app:case_studies} presents the complementary strategies: simplification removes a reasoning requirement, while difficulty preservation retains the target skill under changed conditions. Together with the audit, these examples characterize how the generator adapts problem structure to solver feedback.

\subsubsection{Sensitivity to the labeling model}
\label{sec:labeler_sensitivity}
Solver feedback guides generator RL, while an external reasoning model provides answer labels for downstream solver training. To assess sensitivity to the labeling model, we relabel the same balanced subset of problems from different generators using o3-mini, DeepSeek Pro, and DeepSeek Flash, while holding the filtering rules fixed. We measure pairwise agreement on retention decisions and normalized final answers, considering answer agreement only when both labelers return an extractable answer.

\begin{table}[!t]
\centering
\footnotesize
\setlength{\tabcolsep}{4pt}
\renewcommand{\arraystretch}{1.1}
\caption{Agreement between Labeling Models. All values are percentages on the same generated problems. Retention agreement compares decisions to retain or discard a problem; answer agreement compares normalized final answers.}
\label{tab:labeler_sensitivity}
\begin{tabular}{lrr}
\toprule
Labeler pair & \shortstack{Retention\\agreement} & \shortstack{Answer\\agreement} \\
\midrule
o3-mini vs. DeepSeek Pro & 80.0 & 71.9 \\
o3-mini vs. DeepSeek Flash & 79.3 & 74.5 \\
DeepSeek Pro vs. DeepSeek Flash & 87.3 & 91.8 \\
\bottomrule
\end{tabular}
\end{table}

Retention agreement ranges from 79.3\% to 87.3\%, while answer agreement ranges from 71.9\% to 91.8\%. DeepSeek Pro and DeepSeek Flash show the highest agreement on both measures. These differences highlight the importance of using a shared labeling model and consistent filtering procedures for downstream comparisons.

\begin{table*}[!t]
\centering
\caption{Generalization across Different Seed Sets. ``Seed Set'' trains the solver on the original sampled problems only. ``Augmentation'' expands the seed set with synthetic problems generated by our MATH-trained generator. We report pass@1 accuracy (\%) on the seven mathematical reasoning benchmarks.}
\label{tab:seed}
\resizebox{0.90\textwidth}{!}{%
\footnotesize
\setlength{\tabcolsep}{3.5pt}
\renewcommand{\arraystretch}{1.12}
\begin{tabular}{
    l
    *{7}{S[table-format=2.2]}
    c
}
\toprule
 & {\textbf{MATH}} & {\textbf{GSM8K}} & {\textbf{AMC}} & {\textbf{Minerva}} & {\textbf{Olympiad}} & {\textbf{AIME24}} & {\textbf{AIME25}} & {\textbf{Avg.}} \\
\midrule
\rowcolor{gray!9}
\multicolumn{9}{c}{\textit{\textbf{Qwen3-4B-Base}}} \\
\midrule
MATH Seed Set & 79.60 & 92.79 & 56.35 & 31.73 & 39.76 & 14.52 & 12.02 & 46.68 \\
MATH Augmentation & 82.00 & 92.87 & 60.52 & 32.84 & 44.36 & 17.44 & 14.65 & 49.24 {\scriptsize(+2.56 pp)}\\
\midrule
GSM8K Seed Set & 79.60 & 92.72 & 54.93 & 32.47 & 38.28 & 13.16 & 11.03 & 46.03 \\
GSM8K Augmentation & 78.40 & 93.55 & 59.38 & 32.10 & 41.84 & 13.14 & 13.71 & 47.45 {\scriptsize(+1.42 pp)}\\
\midrule
DAPO Seed Set & 76.00 & 91.88 & 53.82 & 29.52 & 36.94 & 9.78 & 6.80 & 43.53 \\
DAPO Augmentation & 78.20 & 92.19 & 56.75 & 30.26 & 38.58 & 11.94 & 11.63 & 45.65 {\scriptsize(+2.12 pp)}\\
\bottomrule
\end{tabular}%
}
\end{table*}

\subsubsection{Case study: increasing reasoning difficulty}
\label{sec:case_study}
To illustrate the structural adaptation measured in the quality audit, we compare CoT-Self-Instruct~\cite{yu2025cot} with our final model on a high-feedback seed that calls for increased difficulty (Fig.~\ref{fig:case_study_lines_full}). The complementary simplification and difficulty-preservation cases appear in Appendix~\ref{app:case_studies}.

\noindent\textbf{Problem design and solver feedback.}
The generator adds a reasoning step within the same mathematical topic. Together with Appendix~\ref{app:case_studies}, this case illustrates how problem-design reasoning specifies the transformation and solver feedback guides its difficulty. Repeated passages, formatting tags, and non-English text are omitted; condensed reasoning is marked.

\subsection{Co-evolution and Cross-Seed Generalization}
\label{sec:coevolution_generalization}
Finally, we test adaptation to an improving solver and to different seed distributions.
\subsubsection{Generator--solver co-evolution}
\begin{figure}[!t]
    \centering
    \includegraphics[width=0.88\linewidth]{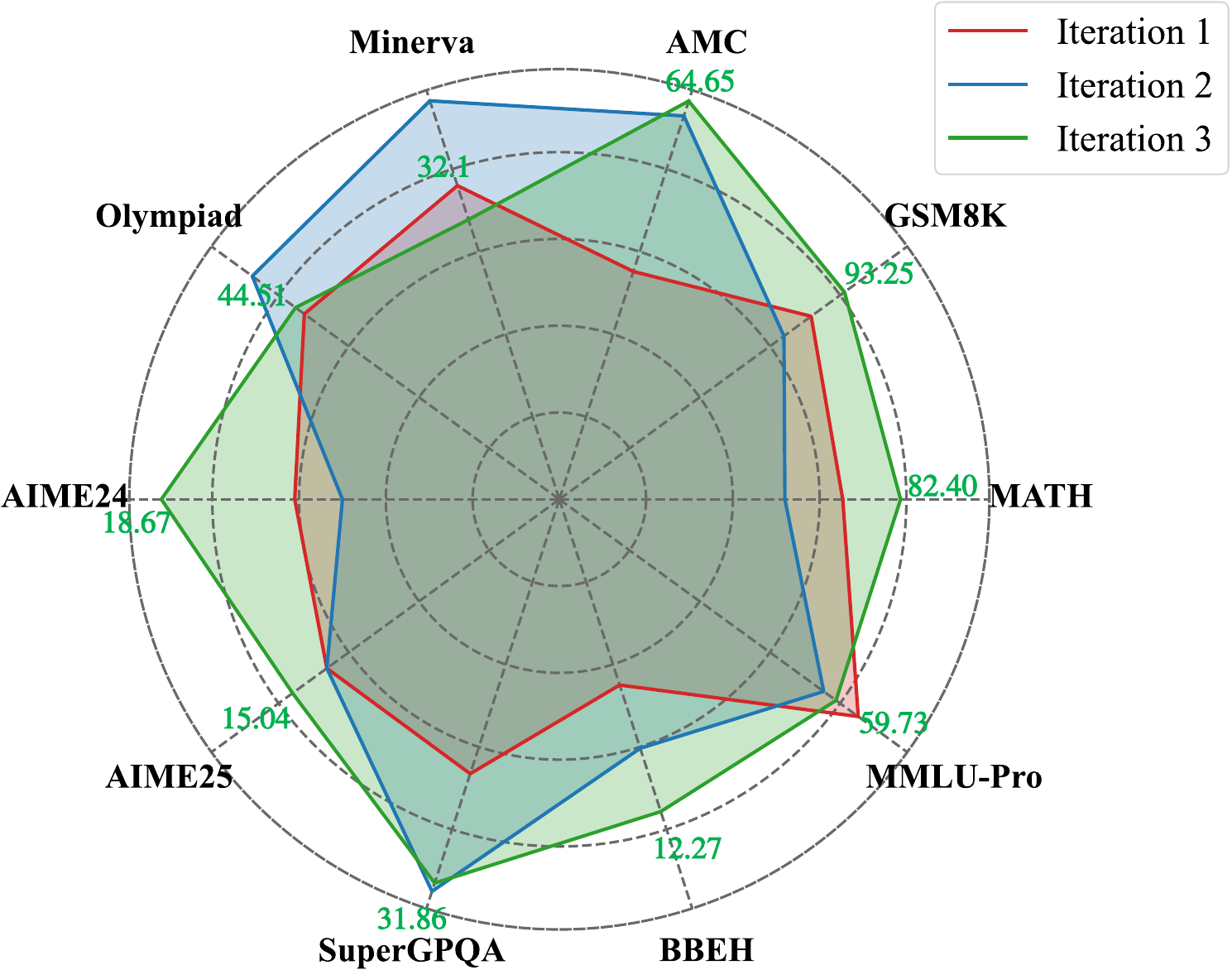}
    \caption{{\bf Performance comparison across iterations on all ten benchmarks.} The solver improves as the generator and solver are trained alternately.}
    \label{fig:radar_iteration}
\end{figure}

We train the problem generator using solver feedback as the reward signal, encouraging the generator’s reasoning on open-ended tasks, while the solver improves its mathematical ability using the generator’s synthetic problems. Both models are initially identical. After each training round, the solver’s enhanced ability alters its difficulty feedback, which we then use to retrain the generator. The updated generator produces more challenging problems, further advancing the solver. Because the reward is computed from the current solver, the generator can adapt as the solver changes across iterations. As shown in Fig.~\ref{fig:radar_iteration}, two additional iterations of this process yield a further 0.70 percentage-point improvement for the solver across ten datasets. These results indicate that the generator and solver can co-evolve, demonstrating the method’s extensibility and the effectiveness of a reward function grounded in solver feedback.

\Cref{tab:coevol} reports the corresponding scores on all ten benchmarks. Average accuracy increases from 44.75\% after the first round to 45.33\% after the second and 45.45\% after the third. The later rounds improve AMC from 60.52\% to 64.65\% and AIME24 from 17.44\% to 18.67\%, while BBEH rises from 11.35\% to 12.27\%. These results extend the first-round gains through repeated interaction between the generator and solver. The full per-benchmark scores are given in Appendix~\ref{app:coevol}.

\subsubsection{Generalization across seed sets}
Our problem generator is trained on the MATH training set. It augments a given seed set by synthesizing additional problems. To evaluate its generalization ability to other seed sets, we also sample 4,000 problems from the training sets of GSM8K and DAPO~\cite{yu2025dapo}, and use our generator to synthesize new problems. The results, reported in Tab.~\ref{tab:seed}, demonstrate the effectiveness of our approach. Our generator exhibits strong generalization across different seed sets, with the synthesized data consistently improving the performance of the original seed sets.

The average gains are 2.56, 1.42, and 2.12 percentage points for the MATH, GSM8K, and DAPO seed sets, respectively. Improvements also extend beyond the benchmark corresponding to the seed source. With GSM8K seeds, augmentation raises AMC accuracy from 54.93\% to 59.38\% and Olympiad accuracy from 38.28\% to 41.84\%. With DAPO seeds, it raises AMC from 53.82\% to 56.75\% and AIME25 from 6.80\% to 11.63\%. Since the generator is trained on MATH in all three settings, these results show that the learned problem-design policy can transform seed problems from other distributions into useful training data without retraining the generator on each seed set.

\FloatBarrier
\section{Conclusion}
Data synthesis is a promising alternative to limited human-curated datasets, but existing methods face two challenges: (i) indiscriminate generation that ignores the solver’s ability and yields low‑value problems; and (ii) a lack of reasoning in problem generation, leading to shallow problem variants. We address these limitations by developing a reasoning‑driven, solver‑adaptive problem generator that extends explicit reasoning beyond problem solving to problem generation. This generator can produce synthetic data to augment existing datasets. We leverage solver feedback on synthesized problems as a reward signal to further strengthen the generator. Comprehensive experiments across mathematical and general reasoning benchmarks show substantial improvements, and ablation studies confirm the contribution of each component.

\begingroup
\microtypesetup{protrusion=false}
\IEEEtriggeratref{58}
\IEEEtriggercmd{\newpage}

\endgroup

\clearpage
\onecolumn
\appendices

\section{Implementation Details}
\label{app:detail}
We train the problem generator with reinforcement learning on the MATH training set, and randomly sample 4,000 problems as a seed set for subsequent synthesis. The VLM generator is trained on the MMK12 training set. All seed-conditioned mathematical synthesis methods use the same 4,000 seed problems, a shared labeling model, and identical filtering and downstream solver-training procedures.

\noindent\textbf{Generator training.}
During SFT, we use a learning rate of $1\times10^{-5}$, a batch size of 120, and 1,200 steps. During RL, we use four rollouts, set the KL penalty coefficient to $1\times10^{-3}$, limit the maximum response length to 6,144 tokens, and train for 20 steps. The consistency score is computed from $m=10$ solver responses per problem.

\noindent\textbf{Synthesis and labeling.}
The trained generator augments each seed once using the prompts in Figs.~\ref{fig:Self-Instruct} and~\ref{fig:our_template}, yielding 4,000 candidate problems. We extract the content within the \texttt{<question>} tags as the problem text. Parsing and solvability filtering lead to different numbers of retained problems across methods. Answers are labeled with o3-mini. In the main results, every problem that passes these filters is combined with the original seed set for solver training. The fixed-size comparison continues sampling from the same seeds until each method retains 4,000 synthetic~problems.

\noindent\textbf{Solver training.}
The problem solver is trained with GRPO using rule-based, verifiable rewards. The solver must place its final answer in the \verb|\boxed| format, using the prompt in Fig.~\ref{fig:solver}. We use AdamW with a cosine learning rate schedule and a peak learning rate of $1\times10^{-6}$, and train for 100 steps with a batch size of 128. We generate eight rollouts per prompt with temperature 1.0, top-p 0.99, and a maximum sequence length of 4,096 tokens. The KL penalty coefficient is $1\times10^{-3}$, and gradients are clipped at a maximum norm of 1.0. We use Fully Sharded Data Parallel with full parameter sharding and optional CPU offloading. GRPO experiments use the EasyR1 codebase~\cite{zheng2025easyr1}, and MathRuler~\cite{mathruler} serves as the program verifier.

\makeatletter
\newenvironment{promptfigure}[2]{%
  \def\promptcaption{#1}%
  \def\promptlabel{#2}%
  \par\vspace{8pt}%
  \noindent\begin{minipage}{\linewidth}%
  \def\@captype{figure}%
  \centering
}{%
  \caption{\promptcaption}\label{\promptlabel}%
  \end{minipage}\par
}
\makeatother

\begin{promptfigure}{Self-Instruct prompt template for problem synthesis.}{fig:Self-Instruct}
\begin{tcolorbox}[colback=blue!5!white,
                  colframe=blue!40!black,
                  arc=1.5mm,
                  auto outer arc,
                  fontupper=\footnotesize\raggedright]
Please create a new problem based on: \verb|<question>|\{Seed Question\}\verb|</question>|. Please reason step by step inside \verb|<think>|...\verb|</think>| and output only the final problem inside \verb|<question>|...\verb|</question>|.
\end{tcolorbox}
\end{promptfigure}

\begin{promptfigure}{Solver Feedback prompt template for problem synthesis.}{fig:our_template}
\begin{tcolorbox}[colback=blue!5!white,
                  colframe=blue!40!black,
                  arc=1.5mm,
                  auto outer arc,
                  fontupper=\footnotesize\raggedright]
Please create a novel self-contained problem with appropriate difficulty adjustment based on: \verb|<question>|\{Seed Question\}\verb|</question>| and student's current accuracy rate: \{Accuracy\}. Apply the following difficulty adjustment rules: If accuracy $<$ 0.3 (low): Simplify the problem significantly---reduce complexity or break it into simpler steps. If $0.3 \leq$ accuracy $\leq 0.7$ (medium): Maintain a similar difficulty level. If accuracy $>$ 0.7 (high): Increase difficulty---add complexity, introduce additional constraints, or combine multiple concepts. Please reason step by step inside \verb|<think>|...\verb|</think>| and output only the final problem inside \verb|<question>|...\verb|</question>|.
\end{tcolorbox}
\end{promptfigure}

\begin{promptfigure}{Solver prompt template for problem solving.}{fig:solver}
\begin{tcolorbox}[colback=blue!5!white,
                  colframe=blue!40!black,
                  arc=1.5mm,
                  auto outer arc,
                  fontupper=\footnotesize\raggedright]
Please reason step by step, and put your final answer within \verb|\boxed{}|. \{Question\}.
\end{tcolorbox}
\end{promptfigure}

\section{Co-evolutionary Results}
\label{app:coevol}
Fig.~\ref{fig:radar_iteration} summarizes the ten-benchmark profile after each round of alternating generator and solver updates. \Cref{tab:coevol} gives the score behind every vertex of that chart. Both models begin from the same first-round checkpoint. After a round, the updated solver supplies new consistency feedback, the generator is trained on that feedback, and the resulting synthetic problems are used to train the solver again.

The ten-benchmark average increases from 44.75\% after the first round to 45.33\% after the second and 45.45\% after the third, a further gain of 0.70 percentage points. The later rounds concentrate on the harder mathematical benchmarks: AMC rises from 60.52\% to 64.65\%, AIME24 from 17.44\% to 18.67\%, and AIME25 from 14.65\% to 15.04\%. MATH and GSM8K change by less than one point and remain near 82\% and 93\%. Among the transfer benchmarks, BBEH increases from 11.35\% to 12.27\% and SuperGPQA stays near 32\%, while MMLU-Pro declines slightly from 60.11\% to 59.73\%. The additional rounds therefore extend the mathematical gains visible on the radar chart, rather than moving every benchmark in the same direction.

\begin{table*}[!t]
\centering
\caption{Performance across Co-evolutionary Iterations on Ten Benchmarks. These scores correspond to the radar chart in Fig.~\ref{fig:radar_iteration}. The generator and solver are trained alternately in successive rounds.}
\label{tab:coevol}
\resizebox{0.96\textwidth}{!}{%
\footnotesize
\setlength{\tabcolsep}{3pt}
\renewcommand{\arraystretch}{1.12}
\begin{tabular}{
    l
    *{11}{S[table-format=2.2]}
}
\toprule
 & {\textbf{MATH}} & {\textbf{GSM8K}} & {\textbf{AMC}} & {\textbf{Minerva}} & {\textbf{Olympiad}} & {\textbf{AIME24}} & {\textbf{AIME25}} & {\textbf{SuperGPQA}} & {\textbf{BBEH}} & {\textbf{MMLU-Pro}} & {\textbf{Avg.}} \\
\midrule
\rowcolor{gray!9}
\multicolumn{12}{c}{\textit{\textbf{Qwen3-4B-Base}}} \\
\midrule
Iteration 1 & 82.00 & 92.87 & 60.52 & 32.84 & 44.36 & 17.44 & 14.65 & 31.33 & 11.35 & $\boldsymbol{60.11}$ & {44.75} \\
Iteration 2 & 81.60 & 92.56 & 64.29 & $\boldsymbol{34.69}$ & $\boldsymbol{45.25}$ & 17.00 & 14.65 & $\boldsymbol{31.90}$ & 11.81 & 59.52 & {45.33} \\
Iteration 3 & $\boldsymbol{82.40}$ & $\boldsymbol{93.25}$ & $\boldsymbol{64.65}$ & 32.10 & 44.51 & $\boldsymbol{18.67}$ & $\boldsymbol{15.04}$ & 31.86 & $\boldsymbol{12.27}$ & 59.73 & {\textbf{45.45}} \\
\bottomrule
\end{tabular}%
}
\end{table*}

\makeatletter
\setlength{\@dblfpsep}{12pt}
\makeatother
\FloatBarrier
\section{Additional Case Studies}
\label[appendix]{app:case_studies}
The case in \cref{sec:case_study} uses a high-consistency seed, so the generator is instructed to increase difficulty. The two cases below cover the other two instructions. A consistency below 0.3 asks for simplification, and a consistency between 0.3 and 0.7 asks for the target skill to be preserved. Each case shows the same seed, the CoT-Self-Instruct~\cite{yu2025cot} continuation, and the problem produced by our final model. Repeated passages, formatting tags, and non-English text are omitted.

\subsection{Simplifying a Multi-Step Geometry Problem}
The seed consistency is 0.222, below the simplification threshold of 0.3. Solving the seed requires both medians and the centroid ratio. A useful simplification should remove that extra relation while still asking for a median length.
\begin{casecomparison}{\textbf{Case study under low solver feedback: simplification.} Responses to the triangle-median problem.}{fig:case_study_median}
\textbf{Original problem:} We have a right triangle $\triangle ABC$ where the legs $AB$ and $BC$ have lengths $6$ and $3\sqrt{3}$, respectively. Medians $AM$ and $CN$ meet at point $P$. What is the length of $CP$?\quad(\emph{Solver feedback: 0.222.})

\textbf{Response of CoT-Self-Instruct:}
Given a right triangle $\triangle ABC$ where the legs $AB$ and $BC$ have lengths $6$ and $3\sqrt{3}$, respectively. Medians $AM$ and $CN$ meet at point $P$. If the area of $\triangle ABC$ is $9\sqrt{3}$, what is the length of $CP$?\par

\par\noindent\rule{\linewidth}{0.1pt}

\textbf{Response of our final model:}
\begin{enumerate}[leftmargin=*,itemsep=1pt,topsep=2pt,parsep=0pt]
\item \textbf{Analyze the student's performance:}  The student's accuracy (0.222) is below the 'low' threshold (0.3). This suggests that the concept being tested is not being well understood or applied. $[\ldots]$
\end{enumerate}
\par
\begin{enumerate}[leftmargin=*,itemsep=1pt,topsep=2pt,parsep=0pt,start=2]
\item \textbf{Determine the appropriate action based on accuracy level:}  Given the accuracy is 'low', the most constructive step is to simplify the problem. The goal is to isolate and focus on the single core geometric property that fails for the student---medians in a triangle. The initial problem combines properties of the median and side lengths. To create a simpler problem:
\end{enumerate}
\par
\begin{enumerate}[leftmargin=*,itemsep=1pt,topsep=2pt,parsep=0pt,start=3]
\item \textbf{Formulate the simplified problem:}
\end{enumerate}
\begin{itemize}[leftmargin=*,itemsep=1pt,topsep=2pt,parsep=0pt]
\item Keep the core geometric object: A right triangle, $\triangle ABC$.
\item $[\ldots]$ This isolates the relationship between a median and its adjacent sides. $[\ldots]$
\end{itemize}
\textbf{Generated problem:} In a right triangle ABC, where angle B is 90 degrees, the lengths of AB and BC are 6 and $3\sqrt{3}$ respectively. If AM is a median, find the length of median AM.\par

\end{casecomparison}

In Fig.~\ref{fig:case_study_median}, CoT-Self-Instruct restates the seed and adds an area condition already implied by the side lengths, leaving the solution procedure unchanged. Our final model removes the second median and the intersection point. The seed requires computing $CN=6$ and applying the centroid relation $CP=\tfrac23 CN=4$, whereas the generated problem asks directly for $AM=3\sqrt{19}/2$. The revision thus eliminates the centroid step while retaining the calculation of a median length.

\subsection{Preserving the Target Skill under Changed Conditions}
The seed consistency is 0.667, inside the preserve band. The intended edit therefore keeps the section formula $P=A+t(B-A)$ and changes the given point or the fraction, rather than replacing the interpolation task.
\begin{casecomparison}{\textbf{Case study under medium solver feedback: preservation.} Responses to the line-segment interpolation problem.}{fig:case_study_interpolation}
\textbf{Original problem:} You connect the points $(0,0)$ and $(9,6)$ with a line segment. Starting at $(0,0)$, you move $\frac13$ of the way along the segment. What is the sum of the coordinates of the point you land upon?\quad(\emph{Solver feedback: 0.667.})

\textbf{Response of CoT-Self-Instruct:}
The student has an accuracy rate of 0.67, which is within the medium difficulty range (0.3 \(\leq\) accuracy $\leq$ 0.7). Therefore, I will maintain a similar level of complexity for the problem.\par
\par
\textbf{Generated problem:} You connect the points (0, 0) and (9, 6) with a line segment. Starting at (0, 0), you move $\frac{1}{3}$ of the way along the segment. What is the sum of the coordinates of the point you land upon?\par
\par
(Note: Use the section formula or parametric equations to find the coordinates.)\par

\par\noindent\rule{\linewidth}{0.1pt}

\textbf{Response of our final model:}
\begin{enumerate}[leftmargin=*,itemsep=1pt,topsep=2pt,parsep=0pt]
\item \textbf{Analyze the student's performance:} The student's accuracy rate is 0.67, which falls in the ``medium'' difficulty range as per the given rules. This suggests they generally understand the core concepts but might face some challenges with the specific application.
\item \textbf{Determine the necessary complexity based on performance:} Given the medium-level performance, the problem should neither be significantly simplified nor made more complex. Therefore, a moderate level of challenge is appropriate.
\item \textbf{Apply the chosen strategy:} To maintain this medium-level challenge, we should keep the core mechanics of the problem but alter the starting point to see if the student's proficiency in applying these mechanics is robust enough to handle a different situation. This tests if they can generalize the skills rather than just perform one specific calculation.
\item \textbf{Formulate the new problem:} A good approach is to change the starting point. We can shift the start from (0,0) to a non-origin point like (2,3). To keep the problem approachable, let's scale the endpoint fraction down slightly, for example, from $\frac{1}{3}$ to $\frac{1}{4}$. This creates a new scenario that assesses the same core skill set but from a slightly different perspective. $[\ldots]$
\end{enumerate}
\textbf{Generated problem:} You connect the points (2, 3) and (9, 6) with a line segment. Starting at (2, 3), you move $\frac{1}{4}$ of the way along the segment. What is the sum of the coordinates of the point you land upon?\par

\end{casecomparison}

Fig.~\ref{fig:case_study_interpolation} illustrates how the generator preserves the solution strategy. CoT-Self-Instruct repeats the seed with an explicit solution hint. Our final model changes the starting point and interpolation fraction while retaining the procedure $P=A+t(B-A)$. The coordinate sum is $5$ for the seed and $15/2$ for the generated problem. The nonzero starting point makes the translation term explicit while preserving the core interpolation procedure.

Taken together, the two cases separate a change of procedure from a change of instance. Low consistency leads our model to drop the centroid relation, whereas intermediate consistency keeps the interpolation formula and changes only its inputs. In both settings, CoT-Self-Instruct stays close to the seed: it either adds a redundant condition or repeats the original problem with a hint.

\end{document}